\documentclass[sigconf]{acmart}

\AtBeginDocument{%
  }
\copyrightyear{2024}
\acmYear{2024}
\setcopyright{acmlicensed}\acmConference[CIKM '24]{Proceedings of the 33rd ACM International Conference on Information and Knowledge Management}{October 21--25, 2024}{Boise, ID, USA}
\acmBooktitle{Proceedings of the 33rd ACM International Conference on Information and Knowledge Management (CIKM '24), October 21--25, 2024, Boise, ID, USA}
\acmDOI{10.1145/3627673.3679885}
\acmISBN{979-8-4007-0436-9/24/10}
\usepackage{multirow}
\usepackage{enumitem}
\usepackage{tikz}
\usepackage{balance}
\begin{document}

\title{ChefFusion: Multimodal Foundation Model Integrating Recipe and Food Image Generation}

\author{Peiyu Li}
\authornote{Both authors contributed equally to this research.}
\email{pli9@nd.edu}
\orcid{0009-0003-4281-9794}
\affiliation{%
  \institution{University of Notre Dame}
  \streetaddress{}
  \city{Notre Dame}
   \state{IN}
  \country{USA}
}

\author{Xiaobao Huang}
\authornotemark[1]
\email{xhuang2@nd.edu}
\orcid{0009-0002-1679-3888}
\affiliation{%
  \institution{University of Notre Dame}
  \streetaddress{}
  \city{}
  \state{}
  \country{}
  \city{Notre Dame}
  \state{IN}
  \country{USA}
}

\author{Yijun Tian}
\email{ytian5@nd.edu}
\affiliation{%
  \institution{University of Notre Dame}
  \streetaddress{}
  \city{}
  \state{}
  \country{}
  \city{Notre Dame}
   \state{IN}
  \country{USA}
}

\author{Nitesh V. Chawla}
\email{nchawla@nd.edu}
\affiliation{%
  \institution{University of Notre Dame}
  \streetaddress{}
  \city{}
  \state{}
  \country{}
  \city{Notre Dame}
    \state{IN}
  \country{USA}
}




\begin{abstract}
Significant work has been conducted in the domain of food computing, yet these studies typically focus on single tasks such as t2t (instruction generation from food titles and ingredients), i2t (recipe generation from food images), or t2i (food image generation from recipes). None of these approaches integrate all modalities simultaneously.
To address this gap, we introduce a novel food computing foundation model that achieves true multimodality, encompassing tasks such as t2t, t2i, i2t, it2t, and t2ti. By leveraging large language models (LLMs) and pre-trained image encoder and decoder models, our model can perform a diverse array of food computing-related tasks, including food understanding, food recognition, recipe generation, and food image generation.
Compared to previous models, our foundation model demonstrates a significantly broader range of capabilities and exhibits superior performance, particularly in food image generation and recipe generation tasks. We open-sourced ChefFusion at \href{https://github.com/Peiyu-Georgia-Li/ChefFusion-Multimodal-Foundation-Model-Integrating-Recipe-and-Food-Image-Generation.git}{GitHub}.
\end{abstract}






\maketitle

\section{Introduction}
Given the fundamental role of food in human life, the field of food computing has recently attracted considerable academic interest \cite{reciperec,recipe2vec,hgat}. This growing area of research has led to numerous studies, each typically focusing on a specific task. For instance, some works \cite{h2020recipegpt, bien2020recipenlg} focus on generating instructions from food titles and ingredients, as well as generating ingredients from recipe titles and cooking instructions, which fall under text-to-text (t2t) tasks. Other studies \cite{salvador2019inverse,chhikara2024fire} concentrate on generating recipes based on food images, which belong to image-to-text (i2t) tasks. Additionally, some research \cite{han2020cookgan,pan2020chefgan} contributes to generating food images from recipes, categorized as text-to-image (t2i) tasks. 

\hspace{0.1em}Despite these advancements, no approach has yet combined all these modalities into an integrated system, highlighting a significant gap. Moreover, recent developments in Transformer-based large language models (LLMs) \cite{vaswani2017attention} and diffusion models \cite{rombach2022high} have shown exceptional performance in various vision and language tasks. However, current methods in food computing have not kept pace with these state-of-the-art (SotA) techniques in natural language processing (NLP) and computer vision (CV).

\hspace{0.1em}To address this gap, we present ChefFusion, a novel food computing foundation model that achieves true multimodality, encompassing tasks such as t2t, t2i, i2t, it2t, and t2ti. ChefFusion integrates these SotA models by employing a pretrained Transformer-based LLM \cite{zhang2022opt} for processing and generating recipes, a visual encoder \cite{radford2021learning} for extracting image features, and an image generation model \cite{rombach2022high} for generating food images. This integration enables ChefFusion to perform a diverse array of food computing-related tasks, including food understanding, food recognition, recipe generation, and food image generation (see Figure \ref{fig:case}).


\hspace{0.1em}The contributions of this paper can be summarized as follows:
\begin{enumerate}[leftmargin=*]
    \item To the best of our knowledge, we present the first general food computing foundation model, which demonstrates a wide suite of multimodal capabilities, including food understanding, food recognition, recipe generation, and food image generation.
    \item Our work pioneers the integration of multimodal dialogue capability into the field of food computing. This innovation enhances user interaction and engagement, leading to more user-friendly and intuitive systems for assisting users with cooking tasks.
    \item We perform a comparative analysis of our results with other prominent methods in food computing. Despite the broader scope of our approach, encompassing multimodal capabilities and functionalities, we demonstrate superior performance, particularly in food image generation and recipe generation tasks. 
\end{enumerate}

\vspace{-0.15in}
\section{Related work}

\textbf{Recipe Generation.} Compared to other i2t tasks, generating detailed recipe information or cooking instructions from a food image presents a considerable challenge. To accomplish this, models need to have comprehensive knowledge of food composition, ingredients, and cooking procedures to ensure accuracy. Constrained by limited model capacity and structure, initial attempts in recipe generation relied heavily on information retrieval techniques \cite{wang2008substructure,xie2010hybrid}. More recent approaches employ encoder-decoder architectures in multimodal settings to generate recipes \cite{salvador2019inverse,wang2022learning,chhikara2024fire}. \cite{salvador2019inverse} introduced a framework that uses encoded representations of images and ingredients in the recipe generation process. \cite{wang2022learning} incorporated tree structures into the encoder-decoder process to include structure-level information. \cite{chhikara2024fire} uses images as input to generate titles and ingredients as intermediate representations, which are then used to create complete recipes with an encoder-decoder model. Instead, we leverage a frozen LLM and CLIP image encoder to generate recipes.
\newline
\newline
\hspace{0.1em}\textbf{Food Image Generation.} Most prior work in image-to-text (i2t) tasks assumes that visual categories are well-structured singular objects, such as birds or flowers. In contrast, food images exhibit significant variability in appearance depending on the ingredients, making them more challenging to generate accurately. Recent approaches often rely on Generative Adversarial Networks (GANs) to generate food images, as seen in studies like \cite{wang2019learning,zhu2019r2gan,papadopoulos2019make,han2020cookgan,pan2020chefgan}. For instance, \cite{wang2019learning} and \cite{zhu2019r2gan} use generative neural networks to produce food images as a constraint to enhance cross-modal recipe retrieval, but these methods typically generate only low-resolution images (e.g., 128 × 128 pixels). \cite{han2020cookgan} and \cite{pan2020chefgan} improves on this by generating higher resolution food images (256 × 256 pixels) based on the ingredients. In contrast to these methods, our approach utilizes a diffusion model to generate food images, achieving even higher resolution (512 × 512 pixels).
\vspace{-0.1in}
\section{Methodology}
\begin{figure*}[t]
    \centering
    \includegraphics[width=7in]{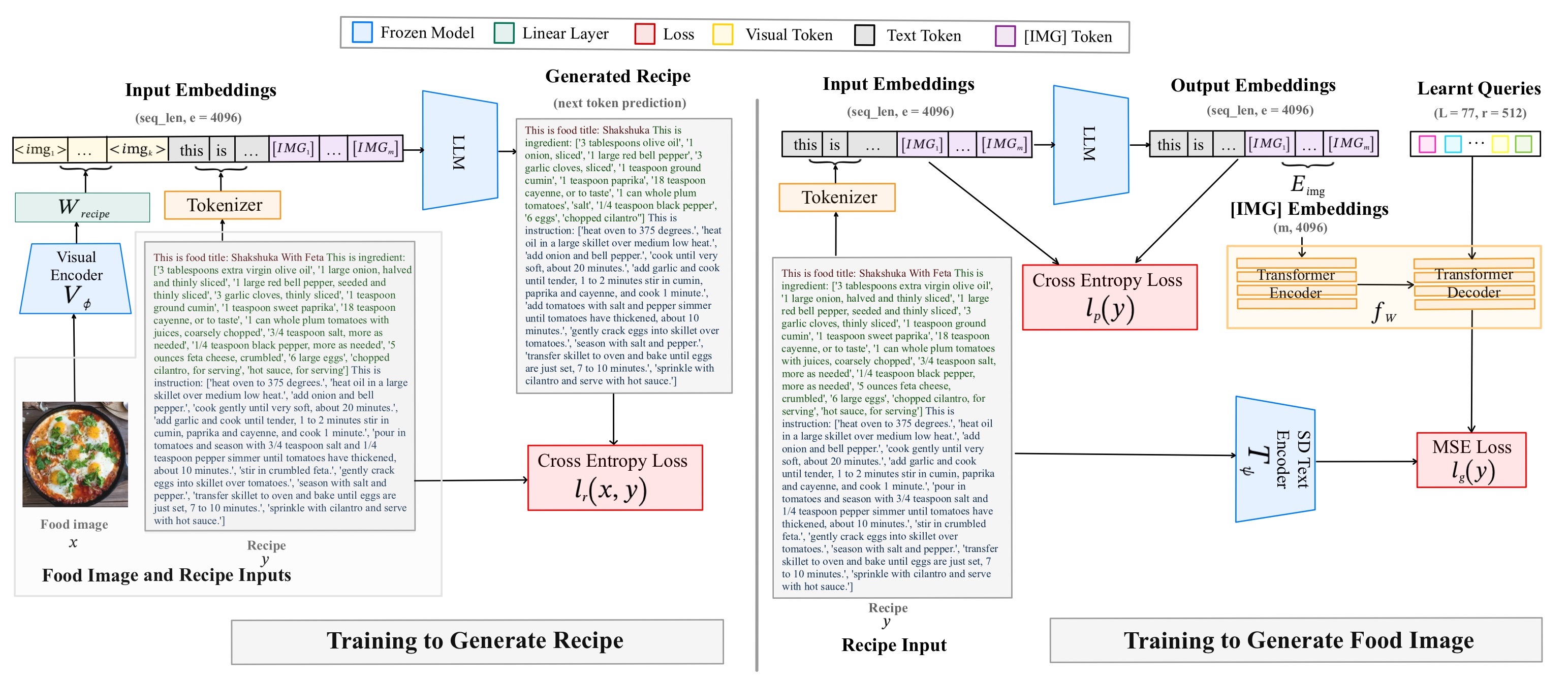}
    \caption{The architecture of ChefFusion: (1) Left: training the model to generate recipe by minimizing $l_{r}(x, y)$; (2) Right: training the model to generate food images by minimizing $l_{g}(y)$ and determine whether to produce text or images at each step by minimizing $l_{p}(y)$.}
    \label{fig:pip-1}
    \Description[<short description>]{<long description>}
    \vspace{-0.20in}
\end{figure*}

The training process consists of two primary components: (1) training the model to generate recipe, and (2) training the model to generate food images. Additionally, the model must determine whether to produce text or images at each step. The detailed architecture is illustrated in Figure \ref{fig:pip-1}.

\subsection{Training to Generate Recipe}
Given an image $x$ and its paired recipe $y$ (tokenized as ($t_{1},…,t_{N}$)), our object is to adapt a frozen LLM to handle sequences of interleaved image and text inputs. We follow previous research \cite{tsimpoukelli2021multimodal,eichenberg2021magma,liu2024visual,koh2023grounding,koh2024generating} in learning translation parameters that convert image features into the text embedding space.

\hspace{0.1em}We start by extracting visual embeddings $v_{\varphi}(x)\in R^{d}$ using a pretrained visual backbone, while keeping its weights $\varphi$ and the LLM weights $\theta$ fixed. We then develop a linear mapping $\mathbf{W}_{recipe}\in \mathbb{R}^{d\times ke}$ to transform $v_{\varphi}(x)$ into a sequence of $k$ $e$-dimensional vectors, which serve as inputs to the LLM (see Figure \ref{fig:pip-1}, left). Here, $e$ denotes the LLM's input embedding dimension.

\hspace{0.1em}We train $\mathbf{W}_{recipe}$ on pairs of food image and recipe by minimizing the negative log-likelihood loss of the token sequence $t_{1},…,t_{N}$:
\begin{equation}
\begin{aligned}l_{r}(x,y) = - \sum_{n=1}^{N}\log p_{\theta}(t_{n}\vert v_{\phi}(x)^{T}\mathbf{W}_{recipe}, t_{1},...,t_{n-1})    
\end{aligned}\end{equation}

\subsection{Training to Generate Food Image}
Following a method similar to \cite{zhou2022learning,koh2023grounding,koh2024generating}, we introduce special $[IMG]$ tokens into the LLM's vocabulary to enable the model to produce image outputs. Specifically, we add a trainable matrix $\mathbf{E}_{img} \in \mathbb{R}^{m\times e}$ to the embedding matrix of the frozen LLM, which represents the $m$ $[IMG]$ token embeddings. According to the experiments of \cite{koh2024generating}, as the number of $[IMG]$ tokens increases, generation generally improves since the inputs to LLM are longer and more expressive. Therefore, we use $m=8$ $[IMG]$ tokens to enhance the expressivity of the frozen LLM for novel image generation. Our objective is to train the model to recognize when to generate $[IMG]$ tokens. This is achieved by minimizing the negative log-likelihood of producing the first $[IMG]$ token, conditioned on the previously generated tokens:
\vspace{-0.05in}
\begin{equation}l_{p}(y) = - \log p_{\{\theta\cup\mathbf{E}_{img}\}}([IMG_{1}]\vert t_{1},...,t_{n} )\end{equation}
 During training, the $[IMG]$ tokens are appended to each recipe. During inference, whenever the first $[IMG_{1}]$ token is generated, the subsequent $m-1$ $[IMG]$ tokens are always produced. 

\hspace{0.1em}To enable our LLM to generate image outputs, the $[IMG]$ tokens must be mapped to a semantically meaningful region within the input space of an image generation model $D_{\psi}$. To achieve this, we use a 4-layer encoder-decoder transformer model \cite{vaswani2017attention} $f_{w}$ with trainable weights $w$. The model $f_{w}$ is conditioned on $h_{\{\theta\cup\mathbf{E}_{img}\}}(y, [IMG])$ and $L$ learned query embeddings $(q_{1},...,q_{L}) \in \mathbb{R}^{L\times r1}$, where $L$ is the maximum input sequence length of the text-to-image generation backbone $D_{\psi}$. We optimize the trainable weights ($(q_{1},...,q_{L})$ and $w$) by minimizing the MSE loss of the model $f_{w}$ outputs against the embeddings produced by the text encoder $T_{\psi}$ of a frozen text-to-image generation model: 
\begin{equation}
\begin{aligned}
l_{g}(y) = \parallel & f_{w}(h_{\{\theta\cup\mathbf{E}_{img}\}}(y,[IMG_{1}]),..., h_{\{\theta\cup\mathbf{E}_{img}\}}(y,[IMG_{m}]), \\
&q_{1},...,q_{L})-T_{\psi}(y)\parallel^{2} 
\end{aligned}
\end{equation}
During inference, when $[IMG]$ tokens are generated, we can synthesize an image:
\begin{equation}
\begin{aligned}
Generated Food Image =& D_{\psi} (f_{w}(h_{\{\theta\cup\mathbf{E}_{img}\}}(y,[IMG_{1}]),..., \\
&h_{\{\theta\cup\mathbf{E}_{img}\}}(y,[IMG_{m}]),q_{1},...,q_{L}))
\end{aligned}\end{equation}

\vspace{-0.1in}
\begin{figure}[t]
    \includegraphics[width=3in]{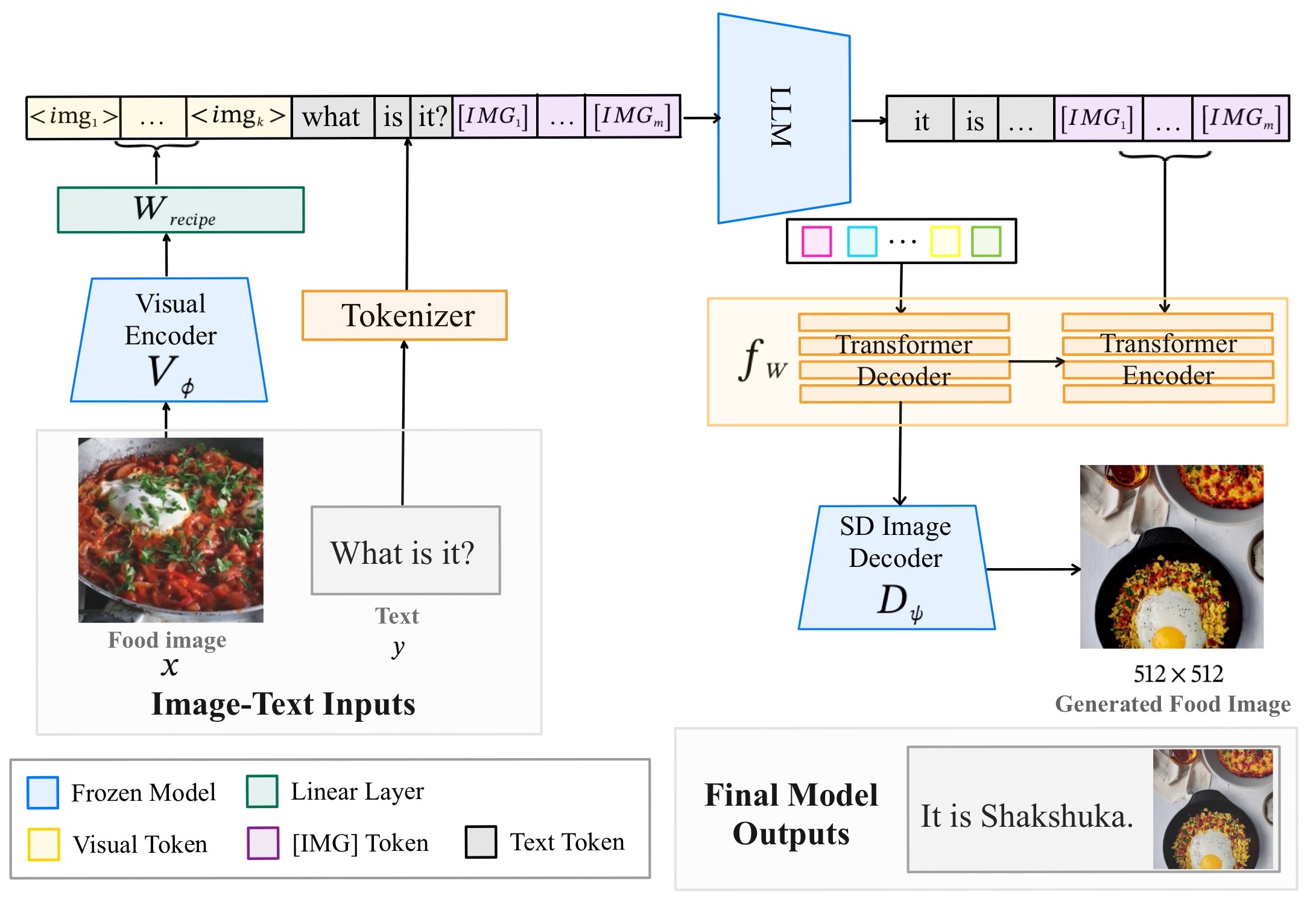}
    \caption{Inference procedure for ChefFusion: The model takes in image and text inputs, and generate text interleaved with food image.}
    \label{fig:pip-2}
    \Description[<short description>]{<long description>}
   \vspace{-0.1in}
\end{figure}

\begin{figure*}
    \centering
    \includegraphics[width=7in]{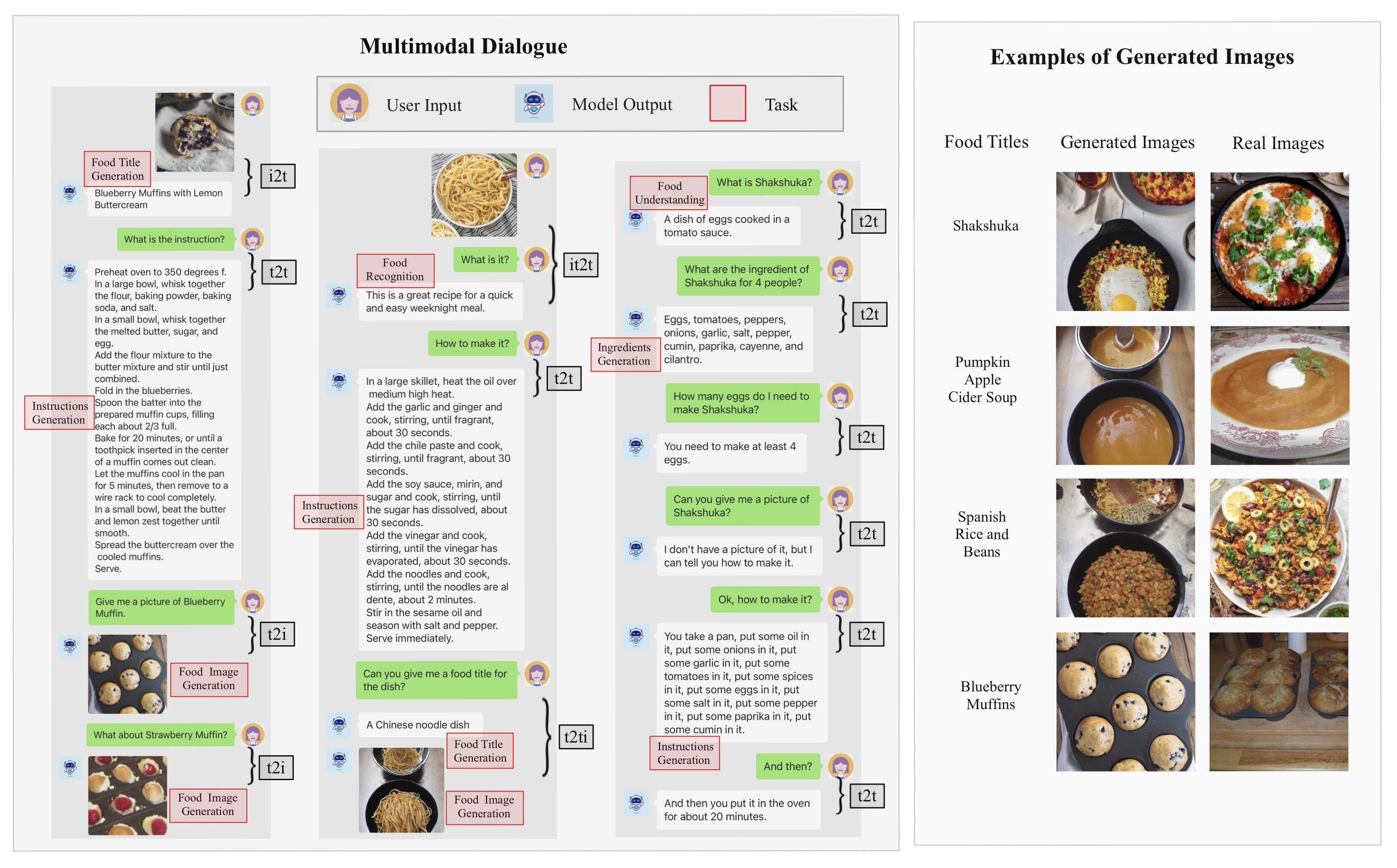}
    \caption{Case Study: ChefFusion demonstrates a wide suite of multimodal capabilities, including food understanding, food recognition, recipe generation, food image generation and multimodal dialogue (left). Example of food images generated by ChefFusion (right).}
    \label{fig:case}
    \Description[<short description>]{<long description>}
    \vspace{-0.1in}
\end{figure*}
\subsection{Dataset and Implement Details}
We train on Recipe1M \cite{salvador2017learning}, which contains more than 1 million recipes and almost 900k images. We use the OPT-6.7B \cite{zhang2022opt} model as the LLM backbone (which produce hidden states $h_{\theta}$ with embedding dim $e$ = 4096). For the visual model used to extract features $v_{\varphi}$, we use the CLIP \cite{radford2021learning} ViT-L model. For our text-to-image generation backbone $D_{\psi}$, we use the Stable Diffusion \cite{rombach2022high} v1.5 model (with $L$ = 77 input vectors).We use $k$ = 4 visual tokens, and $m$ = 8 learnt $[IMG]$ tokens. We set the query embedding dimension $r$ = 512. All pretrained model weights are kept frozen, and we only train the linear layers $\mathbf{W}_{recipe}$, the embedding matrix $\mathbf{E}_{img}$, the parameter $w$ and query vectors $q_{1},...,q_{L}$. We use bfloat16 precision \cite{abadi2016tensorflow}, and optimize using Adam \cite{kingma2014adam} ($\beta_{1}$= 0.9, $\beta_{2}$= 0.95) with a learning rate of 0.001. We train with a batch size of 16 for 14K iterations, which takes 1 day on 2 A100 GPUs.

\section{Experiments}
Our model is a multimodal food foundation model capable of performing text-to-text (t2t), text-to-image (t2i), image-to-text (i2t), image-and-text-to-text (it2t), and text-to-text-and-image (t2ti) tasks. We focus on the most important two evaluation tasks in food computing, i2t (recipe generation) and t2i (food image generation). Other modalities could be found in our case study, shown in Figure \ref{fig:case}. Our results show that our model improves over CookGAN \cite{han2020cookgan}, Stable Diffusion \cite{rombach2022high} and GLIDE \cite{nichol2021glide} in the food image generation task. In the task of food image to recipe task, our model also outperforms the baselines (RecipeNLG \cite{bien2020recipenlg} and InverseCooking \cite{salvador2019inverse}).
\begin{table}[htbp]
\centering
\begin{tabular}{c|c|c}
\toprule
\textbf{Model} & \textbf{SacreBLEU} & \textbf{ROUGE-2}  \\
\midrule
RecipeNLG \cite{bien2020recipenlg}               & 5.03      &0.12     \\ 
InverseCooking \cite{salvador2019inverse}              & 4.27    & 0.11        \\\hline
ChefFusion (Ours)              & \textbf{6.97}      &\textbf{0.12}       \\
\bottomrule
\end{tabular}
\caption{Comparison of Models with different parameters, tuning methods under BLEU and ROUGE metrics}
\vspace{-0.3in}
\label{tab:comparison_i2t}

\end{table}
\vspace{-0.16in}
\begin{table}[htbp]
\centering
\begin{tabular}{c|c}
\toprule
\textbf{Model} & \textbf{CLIP Similarity}   \\
\midrule
GILDE  \cite{nichol2021glide}             & 0.48           \\ 
Stable Diffusion \cite{rombach2022high}              & 0.71            \\
CookGAN \cite{han2020cookgan}             & 0.54            \\   \hline
ChefFusion (Ours)              & \textbf{0.74}             \\
\bottomrule
\end{tabular}
\caption{Comparison of Models with different parameters, tuning methods under BLEU and ROUGE metrics}
\vspace{-0.45in}
\label{tab:comparison_t2i}

\end{table}
\newline
\subsection{Evaluation Metrics}
\textbf{CLIP Similarity}: We utilize the CLIP ViT-L image encoder \cite{CLIP} to generate pooled representations of both generated and real images. Subsequently, we evaluate their cosine similarity, where a higher score signifies a closer resemblance between the generated image and its real counterpart.

\textbf{SacreBLEU}: We use SacreBLEU \cite{sacrebleu} as a reference-based evaluation metric for machine translation. SacreBLEU computes a score based on the n-gram overlap between the machine-generated translations and one or more reference translations. It's commonly used in research and development of machine translation systems to measure their performance against a standard set of reference translations. The higher the SacreBLEU score, the better the translation quality, indicating a higher similarity between the machine-generated translations and the reference translations.

\textbf{ROUGE-2}: We employ ROUGE-2 \cite{rouge} as an evaluation metric that is commonly used in natural language processing. ROUGE-2 evaluates the overlap of bigrams between the generated text and the reference text. It calculates the precision, recall, and F1-score of these bigrams. In essence, ROUGE-2 helps assess how well a machine-generated summary or translation captures the important phrases or concepts present in the reference text at the bigram level. 

\subsection{Tasks}
\textbf{i2t task:} Images in the Recipe1M are utilized as the input for the models and the generated recipes are compared with the ground-truth recipes. In our study, our model shows the best performance both in SacreBLEU and ROUGE-2 of 6.97 and 0.12 respectively compared to the rest of the baseline models, see Table \ref{tab:comparison_i2t}. This indicates that the generated recipes closely resemble human-generated references, implying a high degree of translation accuracy. The performance of our model could be attributed to various factors. Firstly, it leverages LLM and CLIP models as its backbone, enabling it to capture intricate relationships between food images and corresponding recipes more effectively. Secondly, the model may have been trained on a larger and more diverse food dataset, facilitating better generalization to unseen food examples. Furthermore, meticulous hyperparameter tuning and optimization strategies could have contributed to its superior performance.

\textbf{t2i task:} The recipes in the Recipe1M dataset are used as the input for the models and the generated images are compared with the ground-truth images. In Table \ref{tab:comparison_t2i}, our model shows the best performance 0.74 compared to the rest of the models. This suggests that the images generated by this model exhibit a strong alignment with the provided textual descriptions, indicating high fidelity and relevance. To be specific about exceeding the performance of Stable Diffusion, our model enhanced the semantic capturing capability by introducing trainable matrix $\mathbf{E}_{img}$, which enables the CLIP model in our backbone to capture more accurate and relevant information within the recipe context. 
\section{Conclusion}
In this study, we introduce a novel multimodal food computing foundation model that integrates a Transformer-based LLM for recipes, a visual encoder for image features, and an image generation model. This model excels in diverse tasks such as food understanding, recognition, recipe generation, and image generation. Despite the broader scope of our approach, encompassing multimodal capabilities and functionalities, we demonstrate superior performance, particularly in food image generation and recipe generation tasks.


\newpage

\bibliographystyle{ACM-Reference-Format}
\balance 
\bibliography{Reference}

\end{document}